\documentstyle[twocolumn]{article}

\title{Analysis of Japanese Compound Nouns\\using Collocational
Information\thanks{This paper was presented at COLING'94 at Kyoto, August
1994}}
\author{
  \begin{tabular}{ccc}
    KOBAYASI Yosiyuki\\
    TOKUNAGA Takenobu\\
    TANAKA Hozumi
  \end{tabular}\\
  Department of Computer Science\\
  Tokyo Institute of Technology\\
  {\tt \{yashi,take,tanaka\}@cs.titech.ac.jp}}
\date{August}
\def\nl{\hfill\break}
\def\ml#1#2#3{\multicolumn{#1}{#2}{#3}}
\makeatletter
\def\figcap#1#2{\stepcounter{figure}%
  \def\@currentlabel{\thefigure}\label{#2}%
  \par\nobreak\medskip\nobreak{{\normalsize{\bf Fig.~\@currentlabel}\quad#1}}}
\def\tblcap#1#2{\stepcounter{table}%
  \def\@currentlabel{\the\c@table}\label{#2}%
  {\bf Table~\@currentlabel}\quad#1\par\nobreak\medskip\nobreak}
\makeatother

\begin{document}
\maketitle
\section*{Abstract}

Analyzing compound nouns is one of the crucial issues for natural
language processing systems, in particular for those systems that aim
at a wide coverage of domains.  In this paper, we propose a method to
analyze structures of Japanese compound nouns by using both word
collocations statistics and a thesaurus. An experiment is conducted
with 160,000 word collocations to analyze compound nouns of with an
average length of 4.9 characters. The accuracy of this method is about
80\%.

\section{Introduction}
\label{sec:Introduction}

Analyzing compound nouns is one of the crucial issues for natural
language processing systems, in particular for those systems that aim
at a wide coverage of domains. Registering all compound nouns in a
dictionary is an impractical approach, since we can
create a new compound noun by combining nouns.  Therefore, a mechanism
to analyze the structure of a compound noun from the individual nouns is
necessary.

In order to identify structures of a compound noun, we must first find
a set of words that compose the compound noun. This task is trivial
for languages such as English, where words are separated by spaces.
The situation is worse, however, in Japanese where no spaces are
placed between words. The process to identify word boundaries is
usually called segmentation.  In processing languages such as
Japanese, ambiguities in segmentation should be resolved at the same
time as analyzing structure. For instance, the Japanese compound noun
``SinGataKansetuZei''(new indirect tax), produces $16(=2^4)$ segementations
possibilities for this case.  (By consulting a Japanese dictionary, we
would filter out some.) In this case, we have two remaining
possibilities: ``Sin (new)/Gata (type)/Kansetu (indirect)/Zei(tax)'' and
``SinGata (new)/Kansetu (indirect)/ Zei (tax).''\footnote{Here ``/'' denotes a
boundary of words.} We must choose the correct segmentation, ``SinGata
(new)/Kansetu (indirect)/Zei (tax)'' and analyze structure.

Segmentation of Japanese is
difficult only when using syntactic knowledge. Therefore, we could not
always expect a sequence of correctly segmented words as an input to
structure analysis.  The information of structures is also
expected to improve segmentation accuracy.

There are several researches that are attacking this problem. Fuzisaki
{\em et al.\/} applied the HMM model to segmentation and probabilistic
CFG to analyzing the structure of compound nouns~\cite{fuji:91}. The
accuracy of their method is 73\% in identifying correct structures of
kanzi character sequences with average length is 4.2 characters.  In
their approach, word boundaries are identified through purely
statistical information (the HMM model) without regarding such linguistic
knowledge, as dictionaries. Therefore, the HMM model may suggest
an improper character sequence as a word. Furthermore, since
nonterminal symbols of CFG are derived from a statistical analysis of
word collocations, their number tends to be large and so the number of
CFG rules are also large. They assumed compound nouns consist of only one
character words and two character words. It is questionable whether this
method can be extended to handle cases that include more
than two character words without lowering accuracy.

In this paper, we propose a method to analyze structures of Japanese
compound nouns by using word collocational information and a
thesaurus. The collocational information is acquired from a corpus of
four kanzi character words. The outline of procedures to acquire the
collocational information is as follows:
\begin{itemize}
\item extract collocations of nouns from a corpus of
  four kanzi character words
\item replace each noun in the collocations with thesaurus categories,
  to obtain the collocations of thesaurus categories
\item count occurrence frequencies for each collocational pattern
  of thesaurus categories
\end{itemize}
For each possible structure of a compound noun, the preference is
calculated based on this collocational information and the structure
with the highest score wins.

Hindle and Rooth also used collocational information to solve
ambiguities of pp-attachment in English~\cite{hi:91}. Ambiguities are
resolved by comparing the strength of associativity between a
preposition and a verb and the preposition and a
nominal head. The strength of associativity is calculated on the basis
of occurrence frequencies of word collocations in a corpus.  Besides
the word collocations information, we also use semantic knowledge,
namely, a thesaurus.

The structure of this paper is as follows: Section~\ref{sec:CL-info}
explains the knowledge for structure analysis of compound nouns and
the procedures to acquire it from a corpus,
Section~\ref{sec:Algorithm} describes the analysis algorithm, and
Section~\ref{sec:Experiments} describes the experiments that are
conducted to evaluate the performance of our method, and
Section~\ref{sec:Concluding-Remarks} summarizes the paper and discusses
future research directions.

\section{Collocational Information for Analyzing Compound Nouns}
\label{sec:CL-info}

This section describes procedures to acquire collocational
information for analyzing compound nouns from a corpus of four kanzi
character words. What we need is occurrence frequencies of all
word collocations. It is not realistic, however, to collect all
word collocations. We use collocations from thesaurus categories that
are word abstractions.

The procedures consist of the following four steps:
\begin{enumerate}
\item collect four kanzi character words (section~\ref{cwc})
\item divide the above words in the middle to produce pairs of two
kanzi character words; if one is not in the thesaurus, this
four kanzi character word is discarded (section~\ref{cwc})
\item assign thesaurus categories to both two kanzi character word
  (section~\ref{atc})
\item count occurrence frequencies of category  collocations
  (section~\ref{cof})
\end{enumerate}

\subsection{Collecting Word Collocations}
\label{cwc}

We use a corpus of four kanzi character words as the knowledge source of
collocational information.  The reasons are as follows:
\begin{itemize}
\item In Japanese, kanzi character sequences longer than three are
usually compound nouns, This tendency is confirmed by
comparing the occurrence frequencies of kanzi character words in texts
and those headwords in dictionaries.  We investigated the tendency by
using sample texts from newspaper articles and
encyclopedias, and {\em Bunrui Goi Hyou\/} (BGH for short), which is
a standard Japanese thesaurus.  The sample texts include
about 220,000 sentences. We found that three character words and longer
represent 4\% in the thesaurus, but 71\% in the sample texts.
Therefore a collection of four kanzi character words would be used as a
corpus of compound nouns.

\item Four kanzi character sequences are useful to extract binary
  relations of nouns, because dividing a four kanzi character sequence
  in the middle often gives correct segmentation.
  Our preliminary investigation shows that the accuracy of the above
  heuristics is 96 \% (961/1000).

\item There is a fairly large corpus of four kanzi character words
  created by Prof. Tanaka Yasuhito at Aiti Syukutoku college~\cite{TY}.
  The corpus was manually created from newspaper articles and includes
  about 160,000 words.
\end{itemize}

\subsection{Assigning Thesaurus Categories}
\label{atc}

After collecting word collocations, we must assign a thesaurus
category to each word. This is a difficult task because some words are
assigned multiple categories. In such cases, we have several category
collocations from a single word collocation, some of which are
incorrect. The choices are as follows;
\begin{enumerate}
  \def\labelenumi{(\theenumi)}
\item use word collocations with all words is assigned a
  single category.
\item equally distribute frequency of word collcations to all
  possible category collocations~\cite{grish:92}
\item calculate the probability of each category collocation and
distribute frequency based on these probabilities; the probability
of collocations are calculated by using method (2)~\cite{grish:92}
\item determine the correct category collocation by using statistical
  methods other than word collocations~\cite{gut:92,yarowe:92,ver:90,Le:86}
\end{enumerate}

Fortunately, there are few words that are assigned multiple categories
in BGH. Therefore, we use method (1). Word collocations containing
words with multiple categories represent about 1/3 of the corpus. If we used
other thesauruses, which assign multiple categories to more words, we
would need to use method (2), (3), or (4).

\subsection{Counting Occurrence of Category Collocations}
\label{cof}

After assigning the thesaurus categories to words, we count occurrence
frequencies of category collocations as follows:
\begin{enumerate}
\item collect word collocations, at this time we collect only
  patterns of word collocations, but we do not care about occurrence
  frequencies of the patterns
\item replace thesaurus categories with words to produce category
  collocation patterns
\item count the number of category collocation patterns
\end{enumerate}
Note: we do not care about frequencies of word collocations prior to
replacing words with thesaurus categories.

\section{Algorithm}
\label{sec:Algorithm}

The analysis consists of three steps:
\begin{enumerate}
\item enumerate possible segmentations of an input compound noun by
  consulting headwords of the thesaurus (BGH)
\item assign thesaurus categories to all words
\item calculate the preferences of every structure of the compound
  noun according to the frequencies of category collocations
\end{enumerate}

We assume that a structure of a compound noun can be expressed by a
binary tree. We also assume that the category of the right branch of a
(sub)tree represents the category of the (sub)tree itself. This
assumption exsists because Japanese is a head-final language,
a modifier is on the left of its modifiee.
With these assumptions, a preference value of a structure is
calculated by recursive function $p$ as follows:
\begin{displaymath}
  p(t)=\left\{\begin{array}{l}
  1 \mbox{\hskip0.5\columnwidth if $t$ is leaf}\\
  p(l(t))\cdot p(r(t))\cdot cv(cat(l(t)),cat(r(t)))\\
  \mbox{ \hskip0.5\columnwidth otherwise}\\
\end{array}\right.
\end{displaymath}
where function $l$ and $r$ return the left and right subtree of the
tree respectively, $cat$ returns thesaurus categories of the argument.
If the argument of $cat$ is a tree, $cat$ returns the category of the
rightmost leaf of the tree. Function $cv$ returns an associativity
measure of two categories, which is calculated from the frequency of
category collocation described in the previous section. We would use
two measures for $cv$: $P(cat_1, cat_2)$ returns the
relative frequency of collation $cat_1$, which appears on the left side
and $cat_2$, which appears on the right.

\begin{description}
\item[\rm Probability:]\nl
  $\displaystyle cv_1=P(cat_1,cat_2)$
\item[\rm Modified mutual information statistics (MIS):]\nl
  $\displaystyle cv_2=\frac{P(cat_1,cat_2)}{P(cat_1,*)\cdot P(*,cat_2)}$

  where * means don't care.
\end{description}
MIS is similar to mutual infromation used by Church to calculate
semantic dependencies between words~\cite{chu:91}. MIS is different
from mutual information because MIS takes account of the position of
the word (left/right).

Let us consider an example ``SinGataKansetuZei''.
\begin{description}
\item[\rm Segmentation:]two possibilities,\\
  (1) ``SinGata (new)/Kansetu (indirect)/Zei (tax)'' and\\
  (2) ``Sin (new)/Gata (type)/Kansetu (indirect)/Zei (tax)''\\
  remain as mentioned in section~\ref{sec:Introduction}.
\item[\rm Category assignment:]assigning thesaurus categories provides \\
  $(1)'$ ``SinGata [118]/Kansetu [311]/Zei [137]'' and\\
  $(2)'$ ``Sin [316]/Gata [118:141:111]/Kansetu [311]/Zei [137].''\\
  A three-digit number stands for a thesaurus category. A colon ``:''
  separates multiple categories assigned to a word.
\item[\rm Preference calculation:]For the case $(1)'$, the possible
  structures are\\
  {[}[118, 311], 137] and [118, [311, 137]]. \\
  We represent a tree with a list notation.
  For the case $(2)'$, there is an ambiguity with the category ``Sin''
  {[}118:141:111]. We expand the ambiguity to 15 possible structures.
  Preferences are calculated for 17 cases. For example,
  the preference of structure [[118, 311], 137] is calculated as follows:
  \begin{eqnarray*}
    \lefteqn{p([[118,311],137])}\\
    &=&p([118,311])\cdot p(137)\cdot cv(311,137)\\
    &=&p(118)\cdot p(311)\cdot cv(118,311) \cdot cv(311,137)\\
    &=&cv(118,311)\cdot cv(311,137)\\
  \end{eqnarray*}
\end{description}

\section{Experiments}
\label{sec:Experiments}

\subsection{Data and Analysis}

We extract kanzi character sequences from newspaper editorials and
columns and encyclopedia text, which has no overlap with the
training corpus:
954 compound nouns consisting of four kanzi characters, 710 compound
nouns consisting of five kanzi characters, and 786 compound nouns
consisting of six kanzi characters are manually extracted from the set
of the above kanzi character sequences. These three collections of
compound nouns are used for test data.

We use a thesaurus BGH, which is a standard machine
readble Japanese thesaurus. BGH is structured as
a tree with six hierarchical levels. Table~\ref{tab:hie} shows the
number of categories at all levels. In this experiment, we use the
categories at level 3. If we have more compound nouns as knowledge, we
could use a finer hierarchy level.

\begin{center}
  \small
  \tblcap{The number of categories}{tab:hie}
  \begin{tabular}{|l|*{7}{r|}}
    \hline
    Level & 0 & 1 & 2 & 3 & 4 & 5 & 6 \\
    \hline
    No. of Cat. & 1 & 3 & 13 & 94 & 510 & 833 & 6023\\
    \hline
  \end{tabular}
\end{center}

As mentioned in Section~\ref{sec:CL-info}, we create a set of
collocations of thesaurus categories from a corpus of four kanzi
character sequences and BGH.  We analyze the test data according to
the procedures described in Section~\ref{sec:Algorithm}. In
segmentation, we use a heuristic ``minimizing the number of content
words'' in order to prune the search space. This heuristics is
commonly used in the Japanese morphological analysis. The correct
structures of the test data manually created in advance.

\subsection{Results and Discussions}

Table~\ref{tab:rsl1} shows the result of the analysis for four, five, and
six kanzi character sequences. ``$\infty$'' means that the correct
answer was not obtained because the heuristics is segmentation
filtered out from the correct segmentation. The first row shows the
percentage of cases where the correct answer is uniquely
identified, no tie. The rows, denoted ``$\sim n$'', shows
the percentage of correct answers in the n-th rank.
$4 \sim$ shows the percentage of
correct answers ranked lower or equal to 4th place.

\begin{center}
  \tblcap{Accuracy of analysis [\%]}{tab:rsl1}
  \begin{tabular}{|r|*{6}{c|}}
    \hline
    & \ml{2}{c|}{4 kanzi}&\ml{2}{c|}{5 kanzi}&\ml{2}{c|}{6 kanzi}\\
    \cline{2-7}
    rank& $cv_1$& $cv_2$& $cv_1$& $cv_2$& $cv_1$& $cv_2$\\
    \hline
    1 & 96 & 96& 63& 59& 48& 53\\
    $\sim1$& 97& 96& 71& 68& 54& 68\\
    $\sim2$& 99& 99& 91& 91& 89& 93\\
    $\sim3$& 99& 99& 92& 92& 91& 94\\
    \hline
    4$\sim$& 0.1& 0.1& 2& 2& 4& 4\\
    \hline
    $\infty$& 1& 1& 6& 6& 5& 2\\
    \hline
  \end{tabular}
\end{center}

Regardless, more than 90\% of the correct answers are within the second rank.
The probabilistic measure
$cv_1$ provides better accuracy than the mutual information measure
$cv_2$ for five kanzi character compound nouns, but the result is
reversed for six kanzi character compound nouns. The results for four
kanzi character words are almost equal. In order to judge which measure is
better, we need further experiments with longer words.

We could not obtain correct segmentation for 11 out of 954 cases for four
kanzi character words, 39 out of 710 cases for five kanzi character
words, and 15 out of 787 cases for six kanzi character words.
Therefore, the accuracy of segmentation candidates are 99\%(943/954),
94.5\% (671/710) and 98.1\% (772/787) respectively. Segmentation
failure is due to
words missing from the dictionary and the heuristics we adopted.

As mentioned in Section 1, it is difficult to correct segmentation by
using only syntactic knowledge. We used the heuristics to reduce
ambiguities in segmentation, but ambiguities may remain. In
these experiments, there are 75 cases where ambiguities can not be
solved by the heuristics. There are 11 such cases for four kanzi
character words, 35 such cases for five kanzi character words, and 29
cases for six kanzi character words. For such cases, the correct
segmentation can be uniquely identified by applying the structure
analysis for 7, 19, and 17 cases, and the correct structure can be uniquely
identified for 7, 10, and 8 cases for all collections of test data by using
$cv_1$. On the other hand, 4, 18, and 21 cases correctly
segmented and 7, 11, and 8 cases correctly analyzed their structures
for all collections by using $cv_2$.

For a sequence of segmented words, there are several possible
structures. Table~\ref{structure} shows possible structures for
four words sequence and their occurrence in all data
collections. Since a compound noun of our test data consists of four,
five, and six characters, there could be cases with a compound noun
consisting of four, five, or six words. In the current data collections,
however, there are no such cases.

In table~\ref{structure}, we find significant deviation over
occurrences of structures. This deviation has strong correlation with
the distance between modifiers and modifees. The rightmost column
(labeled $\sum d$) shows sums of distances between modifiers and
modifiee contained in the structure. The distance is measured based on
the number of words between a modifier and a modifiee. For instance, the
distance is one, if a modifier and a modifiee are immediately adjacent.

The correlation between the distance and the occurrence of structures
tells us that a modifier tends to modify a closer modifiee. This
tendency has been experimentally proven by
Maruyama~\cite{maruyama:92:a}. The tendency is expressed by the
formula that follows:
\[
  q(d) = 0.54 \cdot d^{-1.896}
\]
where $d$ is the distance between two words and $q(d)$ is the
probability when two words of said distance is $d$ and have a modification
relation.

We redifined $cv$ by taking this tendency as the formula that follows:
\[
  cv' = cv \cdot q(d)
\] where $cv'$ is redifined $cv$. Table~~\ref{tab:rsl1} shows the result
by using new $cv$s. We obtained significant improvement in 5 kanzi and 6
kanzi collection.

\begin{center}
  \tblcap{Table of possible structures}{structure}
  \begin{tabular}{|l|r|r|r|}
    \hline
    structure& 5 kanzi& 6 kanzi& $\sum d$\\
    \hline
    [$w_1$]& 0& 1& 0\\
    \hline
    [$w_1$, $w_2$]& 268& 85& 1\\
    \hline
    {[[$w_1$, $w_2$], $w_3$]}& 269& 405& 2\\
    {[$w_1$, [$w_2$, $w_3$]]}&  96& 166& 3\\
    \hline
    {[[[$w_1$, $w_2$], $w_3$], $w_4$]}& 16& 48& 3\\
    {[[$w_1$, $w_2$], [$w_3$, $w_4$]]}& 13& 45& 4\\
    {[$w_1$, [$w_2$, [$w_3$, $w_4$]]]}&  2&  3& 6\\
    {[$w_1$, [[$w_2$, $w_3$], $w_4$]]}&  3&  8& 5\\
    {[[$w_1$, [$w_2$, $w_3$]], $w_4$]}&  4& 11& 4\\
    \hline
  \end{tabular}
\end{center}

\begin{center}
  \tblcap{Accuracy of analysis [\%]}{tab:rsl2}
  \begin{tabular}{|r|*{6}{c|}}
    \hline
    & \ml{2}{c|}{4 kanzi}&\ml{2}{c|}{5 kanzi}&\ml{2}{c|}{6 kanzi}\\
    \cline{2-7}
    rank& $cv_1$& $cv_2$& $cv_1$& $cv_2$& $cv_1$& $cv_2$\\
    \hline
    1 & 96& 97& 73& 79& 62& 70\\
    $\sim1$& 97& 97& 73& 79& 62& 71\\
    $\sim2$& 99& 99& 91& 92& 90& 94\\
    $\sim3$& 99& 99& 93& 93& 92& 96\\
    \hline
    $\sim4$& 0.1& 0.1& 2& 2& 5& 3\\
    \hline
    $\infty$& 1& 1& 5& 5& 1& 1\\
    \hline
  \end{tabular}
\end{center}

We assumed that the thesaurus category of a tree be represented by the
category of its right branch subtree because Japanese is a head-final
language. However, when a right subtree is a word such as
suffixes, this assumption does not always hold true.  Since our ultimate aim
is to analyze semantic structures of compound nouns, then dealing with only
the grammatical head is not enough. We should take semantic
heads into consideration. In order to do so, however, we need knowledge to
judge which subtree represents the semantic features of the tree. This
knowledge may be extracted from corpora and machine readable
dictionaries.

A certain class of Japanese nouns (called {\em Sahen meisi\/}) may
behave like verbs. Actually, we can make verbs from these nouns by
adding a special verb ``{\em -suru\/}.'' These nouns have case
frames just like ordinary verbs. With compound nouns including such
nouns, we could use case frames and selectional restrictions to
analyze structures. This process would be almost the same as
analyzing ordinary sentences.

\section{Concluding Remarks}
\label{sec:Concluding-Remarks}

We propose a method to analyze Japanese compound nouns using
collocational information and a thesaurus. We also describe a method
to acquire the collocational information from a corpus of four kanzi
character words. The method to acquire collocational information is
dependent on the Japanese character, but the method to calculate
preferences of structures si applicable to any language with compound
nouns.

The experiments show that when the method analyzes compound nouns with an
average length 4.9, it produces an accuracy rate of about 83\%.

We are considering those future works that follow:
\begin{itemize}
\item incorporate other syntactic information, such as affixes knowledge
\item use another semantic information as well as thesauruses, such as
  selectional restriction
\item apply this method to disambiguate other syntactic structures
  such as dependency relations
\end{itemize}

\end{document}